\documentclass[sigconf]{acmart}

\AtBeginDocument{%
  }

\copyrightyear{2026}
\acmYear{2026}
\setcopyright{cc}
\setcctype{by}
\acmConference[SIGIR '26]{Proceedings of the 49th International ACM SIGIR Conference on Research and Development in Information Retrieval}{July 20--24, 2026}{Melbourne, VIC, Australia}
\acmBooktitle{Proceedings of the 49th International ACM SIGIR Conference on Research and Development in Information Retrieval (SIGIR '26), July 20--24, 2026, Melbourne, VIC, Australia}
\acmDOI{10.1145/3805712.3809581}
\acmISBN{979-8-4007-2599-9/2026/07}

\usepackage{multirow}
\usepackage{tabularray}  
\usepackage{amsmath} 
\usepackage{csquotes}  
\usepackage{amsthm}
\usepackage{subcaption} 
\usepackage[htt]{hyphenat} 

\begin{document}

\title[Looking for the Bottleneck in Fine-grained Temporal Relation Classification]{Looking for the Bottleneck in Fine-grained \texorpdfstring{\\}{} Temporal Relation Classification}

\author{Hugo Sousa}
\authornote{Work done before joining Amazon.}
\orcid{0000-0003-3226-9189}
\affiliation{%
  \institution{University of Porto \\ INESC TEC}
  \city{Porto}
  \country{Portugal}
}
\email{hugo.o.sousa@inesctec.pt}

\author{Ricardo Campos}
\orcid{0000-0002-8767-8126}
\affiliation{%
  \institution{University of Beira Interior \\ INESC TEC}
  \city{Porto}
  \country{Portugal}
}
\email{ricardo.campos@inesctec.pt}

\author{Alípio Jorge}
\orcid{0000-0002-5475-1382}
\affiliation{%
  \institution{University of Porto \\ INESC TEC}
  \city{Porto}
  \country{Portugal}
}
\email{alipio.jorge@inesctec.pt}


\begin{abstract}
  Temporal relation classification is the task of determining the temporal relation between pairs of temporal entities in a text.
  Despite recent advancements in natural language processing, temporal relation classification remains a considerable challenge.
  Early attempts framed this task using a comprehensive set of temporal relations between events and temporal expressions.
  However, due to the task complexity, datasets have been progressively simplified, leading recent approaches to focus on the relations between event pairs and to use only a subset of relations.
  In this work, we revisit the broader goal of classifying interval relations between temporal entities by considering the full set of relations that can hold between two time intervals.
  The proposed approach, Interval from Point, involves first classifying the point relations between the endpoints of the temporal entities and then decoding these point relations into an interval relation.
  Evaluation on the TempEval-3 dataset shows that this approach can yield effective results, achieving a temporal awareness score of $70.1$ percent, a new state-of-the-art on this benchmark.
\end{abstract}



\maketitle

\section{Introduction} \label{sec:introduction}

Understanding temporal information in text is essential for comprehending its underlying narrative~\cite{Leeuwenberg2019}.
Many real-world applications, such as summarization~\cite{zhang-etal-2024-cross}, question answering~\cite{chen-etal-2023-multi}, and timeline generation~\cite{alsayyahi-batista-navarro-2023-timeline}, rely on the ability to determine when events occur and how they relate in time.
Consequently, extracting and structuring temporal information is an important task in natural language processing.

A common approach to structuring temporal information is by encoding the relative temporal positioning of temporal entities -- entities that can be anchored in time -- occurring in the text.
This approach has been extensively studied, particularly within the framework of the TimeML annotation scheme~\cite{Saur2006} and the TempEval shared tasks~\cite{Verhagen2007,Verhagen2010,UzZaman2013}.
In this framework, temporal information is structured as triples of the form (source entity, relation, target entity), where the relation belongs to a set of 14 labels inspired by the 13 interval relations defined by~\citet{10.1145/182.358434}.
For instance, in the sentence \textit{\enquote{John \textbf{arrived} in Boston after \textbf{10 p.m.}}}, the temporal entities \textit{\enquote{arrived}} and \textit{\enquote{10 p.m.}} are linked by the \texttt{after} relation, forming the triplet (\textit{\enquote{arrived}}, \texttt{after}, \textit{\enquote{10 p.m.}}).

While temporal relations are intuitive for humans, they are often conveyed implicitly, relying on reasoning and commonsense knowledge~\cite{Han2021,Zhou2019} to be inferred -- abilities that computational models still struggle to replicate~\cite{Hasegawa2024}.
A key obstacle for further development of effective temporal relation classification systems lies in the complexity of creating datasets annotated with the full set of temporal relations.
The process is not only labor-intensive but also hindered by low inter-annotator agreement and data sparsity.
As a result, existing datasets remain limited, typically containing only a few thousand annotated relations~\cite{Pustejovsky2003TimeBank,UzZaman2013,Styler2014}.
Moreover, these annotations end up being highly skewed toward the most frequent relations, leaving the others significantly underrepresented.

To address these challenges, prior research has sought to simplify the task in various ways, including reducing the number of relation types~\cite{Cassidy2014}, focusing exclusively on relations between events~\cite{Naik2019}, restricting annotations to specific event subsets~\cite{alsayyahi-batista-navarro-2023-timeline}, or considering only relations between event start points~\cite{Ning2018}.
While these approaches have advanced our understanding of temporal structures, systems trained on such datasets do not generalize beyond the specific relations and event types encountered during training.
This limitation restricts their applicability in real-world applications.
For instance, models that do not classify relations between events and temporal expressions cannot effectively anchor events in time.
Furthermore, since the introduction of these benchmarks, most research efforts have been dedicated to producing systems for them~\cite{Mathur2021,Zhou2022,Man2022,Wang2023}, leaving the classification of the full set of interval relations largely unexplored.

In this work, we revisit the challenge of classifying fine-grained temporal relations at the interval level, using both events and temporal expressions as entities.
The proposed system, \textbf{Interval from Point} (IfP), decomposes the problem by first classifying the pointwise relations between the endpoints of temporal entities and then decoding these point relations into an interval relation.
The main motivation behind this strategy is that at the point-level, classification involves only three labels -- before, after, and equals -- which simplifies the learning task.
Furthermore, since each relation in the interval relation dataset can be translated to four point relations, this approach significantly increases the number of training examples.
This enables us to fine-tune a language model for point relation classification, yielding effective results at this level.
The predicted point relations are then aggregated to infer the corresponding interval relation through a decoding strategy.
Evaluation on the TempEval-3 dataset shows that this approach can yield effective results and a clear path for future research.

In summary, the contributions of this research are the following:

\textbf{(i)}~We propose a new method for temporal relation classification, IfP, which breaks down the task of interval relation classification into classifying point relations between the endpoints of temporal entities.

\textbf{(ii)}~We systematically evaluate the impact of data augmentation and model size on the effectiveness of point relation classification model.

\textbf{(iii)}~Our experiments demonstrate that fine-tuning a pre-trained language model on an augmented interval dataset can surpass the effectiveness of the best baseline systems from previous studies.

\textbf{(iv)}~Finally, our experimental setup allows us to identify the current limitations of the proposed system and provide a clear path for future research.

\section{Temporal Relation Classification} \label{sec:temporal-relation-classification}

The goal of the temporal relation classification task is to find a system $S$ that, given a document $d$ and a pair of temporal entities $x$ and $y$ within it, determines their temporal relation $r$:

\begin{equation}
    S(d, x, y) \to r
\end{equation}

Regarding the \textbf{temporal entities} $x$ and $y$, these are pre-annotated in the text and can be thought of as any entity that can be anchored in time.
In TimeML framework~\cite{Saur2006}, the most established annotation scheme, temporal entities are defined as being composed of events and temporal expressions.
Events describe situations that occur, unfold, or persist over time, including actions (\enquote{announced}), states (\enquote{remained closed}), and conditions (\enquote{ongoing}). They can appear as verbs, nominalizations, adjectives, or prepositional phrases.
Temporal expressions, on the other hand, denote points or intervals in time and provide explicit temporal anchors within a text.
These can be dates (\enquote{January 1}), times (\enquote{midnight}), durations (\enquote{two years}), or sets (\enquote{every Thursday}).
A temporal expression of particular importance is the document creation time.
Although it is not explicitly mentioned in the text, this temporal expression holds the time and date when the document was created and serves as a reference point for determining whether an entity occurs in the past, present, or future.

\begin{figure}[t]
    \centering
    \includegraphics[width=\columnwidth]{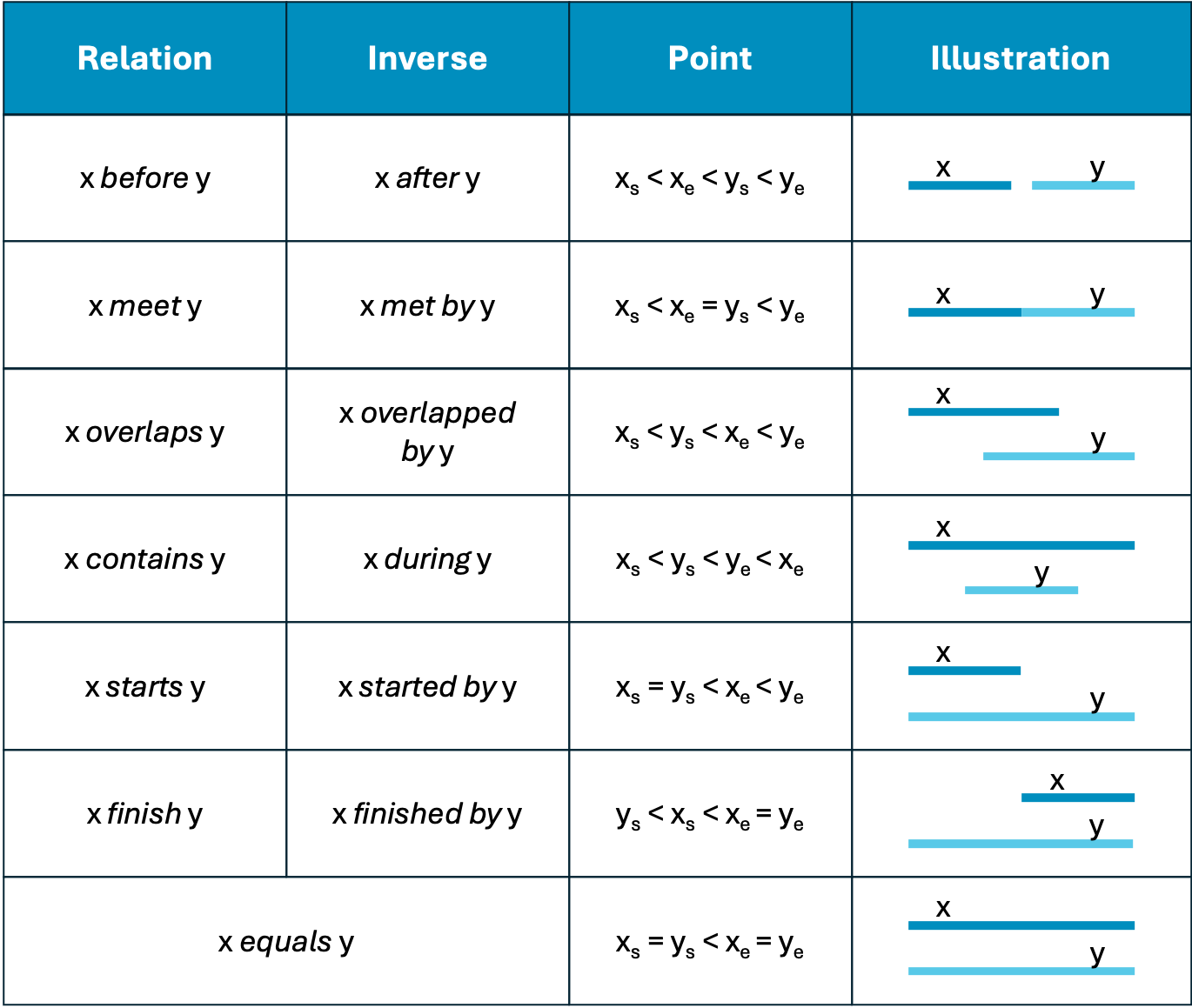}
    \caption{Allen's relations between two time intervals along with the point relations that define them. $x$ and $y$ are the entities, and the subscripts $s$ and $e$ stand for their start and end, respectively.}
    \Description{Table listing the 13 Allen interval relations between two intervals $x$ and $y$ (before, after, meets, met by, overlaps, overlapped by, starts, started by, finishes, finished by, contains, during, equals). Each row shows a timeline diagram of intervals $x$ and $y$ aligned to illustrate the relation, paired with the four point relations between the endpoints $x_s$, $x_e$, $y_s$, $y_e$ that uniquely define it.}
    \label{fig:temporal_relations}
\end{figure}

Regarding the \textbf{temporal relation} $r$, this is an element of the set of 13 temporal relations $R$ that describe how the entities are positioned in time with respect to one another.
This set of temporal relations was first discussed in Allen's seminal work~\cite{10.1145/182.358434}, where the focus was on finding an exhaustive and mutually exclusive set of temporal relations that can hold between two time intervals.
The set of relations are illustrated in Figure~\ref{fig:temporal_relations} and are composed of six fundamental relations, their inverses, and the symmetric \texttt{equals} relation.

It is worth noting that TimeML's set of relations does not exactly match Allen's interval relations.
Specifically, the guidelines do not include labels corresponding to \texttt{overlaps} and \texttt{overlapped-by} relations.
We further discuss this in Section~\ref{ssec:datasets} when we describe the TempEval-3 dataset, which was annotated according to the TimeML guidelines.

\section{Interval from Point System} \label{sec:interval-from-point-system}

One particular aspect of Allen's relations is that they can be described by the point relations between the start and end of the entities involved (see Figure~\ref{fig:temporal_relations}).
Given a temporal entity $x$, we define $x_s$ and $x_e$ as its start and end points, respectively.
The temporal relation between the entities $x$ and $y$ can then be decomposed into four point relations between the endpoint pairs $(x_s,y_s)$, $(x_s,y_e)$, $(x_e,y_s)$, and $(x_e,y_e)$.
For instance, the interval relation \texttt{starts} translates into: $(x_s,\texttt{=},y_s)$, $(x_s,\texttt{<},y_e)$, $(x_e,\texttt{>},y_s)$, and $(x_e,\texttt{<},y_e)$.
Note that at the point level only three relations are possible -- before (\texttt{<}), equals (\texttt{=}), and after (\texttt{>}) -- which we collectively denote as $R_p = \{\texttt{<},\texttt{=},\texttt{>}\}$.

The \textit{Interval from Point} (IfP) system we propose leverages this decomposition to model interval relations by predicting the point relations between the endpoints of the entities.
To achieve this, we train a point relation classification model $P$ that, when given a document $d$ and a pair of entity endpoints $x_i \in \{x_s, x_e\}$ and $y_j \in \{y_s, y_e\}$, predicts the probability distribution over the three possible point relations:

\begin{equation}
    P(d,x_i,y_j) = \begin{bmatrix} \hat{p}_{x_i,y_j,<} \\ \hat{p}_{x_i,y_j,=} \\ \hat{p}_{x_i,y_j,>} \end{bmatrix}
\end{equation}

where $\hat{p}_{x_i, y_j, r_{p}}$ is the estimated probability for the point relation $r_{p} \in R_{p}$ between $x_i$ and $y_j$.

Since each point relation has an inverse, the estimated probability for $r_p$ should be the same as for the inverse relation $r_p^{-1}$ when the source and target are swapped.
For example, going back to the \textit{\enquote{John \textbf{arrived} in Boston after \textbf{10 p.m.}}} example, $\hat{p}_{x_s, y_e,<}$ when $x=\textit{\enquote{arrived}}$ and $y=\textit{\enquote{10 p.m.}}$ should be approximately the same as $\hat{p}_{y_s,x_e,>}$ when $x=\textit{\enquote{10 p.m.}}$ and $y=\textit{\enquote{arrived}}$.

At inference time, our system takes advantage of this property to estimate the probability distribution for the point relations between $x_i$ and $y_j$, $\mathbf{\hat{p}_{ij}}$:

\begin{equation} \label{eq:point-relation-distribution}
    \begin{split}
        \mathbf{\hat{p}_{ij}} & = P(d, x_i, y_j) \odot J \cdot P(d, y_i, x_j)                                                                                                                                         \\
                              & = \begin{bmatrix} \hat{p}_{x_i, y_j,<} \cdot \hat{p}_{y_i, x_j,>} \\ \hat{p}_{x_i, y_j,=} \cdot \hat{p}_{y_i, x_j,=} \\ \hat{p}_{x_i, y_j,>} \cdot \hat{p}_{y_i, x_j,<} \end{bmatrix} = \begin{bmatrix} \hat{p}_{ij,<} \\ \hat{p}_{ij,=} \\ \hat{p}_{ij,>} \end{bmatrix}
    \end{split}
\end{equation}

where $J \in \mathbb{R}^{3 \times 3}$ is the anti-identity matrix (i.e., 1s on the anti-diagonal, 0s elsewhere) and $\odot$ denotes the Hadamard product.
The probability distributions of the four endpoint pairs -- $\mathbf{\hat{p}_{ss}}$, $\mathbf{\hat{p}_{se}}$, $\mathbf{\hat{p}_{es}}$, and $\mathbf{\hat{p}_{ee}}$ -- are then passed to the decoding strategy $D$ to determine the most likely interval relation $\hat{r} \in R$, which we use as the predicted relation between $x$ and $y$:

\begin{equation}
    \textit{IfP}(d, x, y) = \hat{r} =  D(\mathbf{\hat{p}_{ss}}, \mathbf{\hat{p}_{se}}, \mathbf{\hat{p}_{es}}, \mathbf{\hat{p}_{ee}})
\end{equation}

The next sections detail the point classification model and decoding strategy.

\subsection{Point Relation Classification} \label{ssec:point-relation-classification}

\begin{figure}[ht]
    \centering
    \includegraphics[width=\columnwidth]{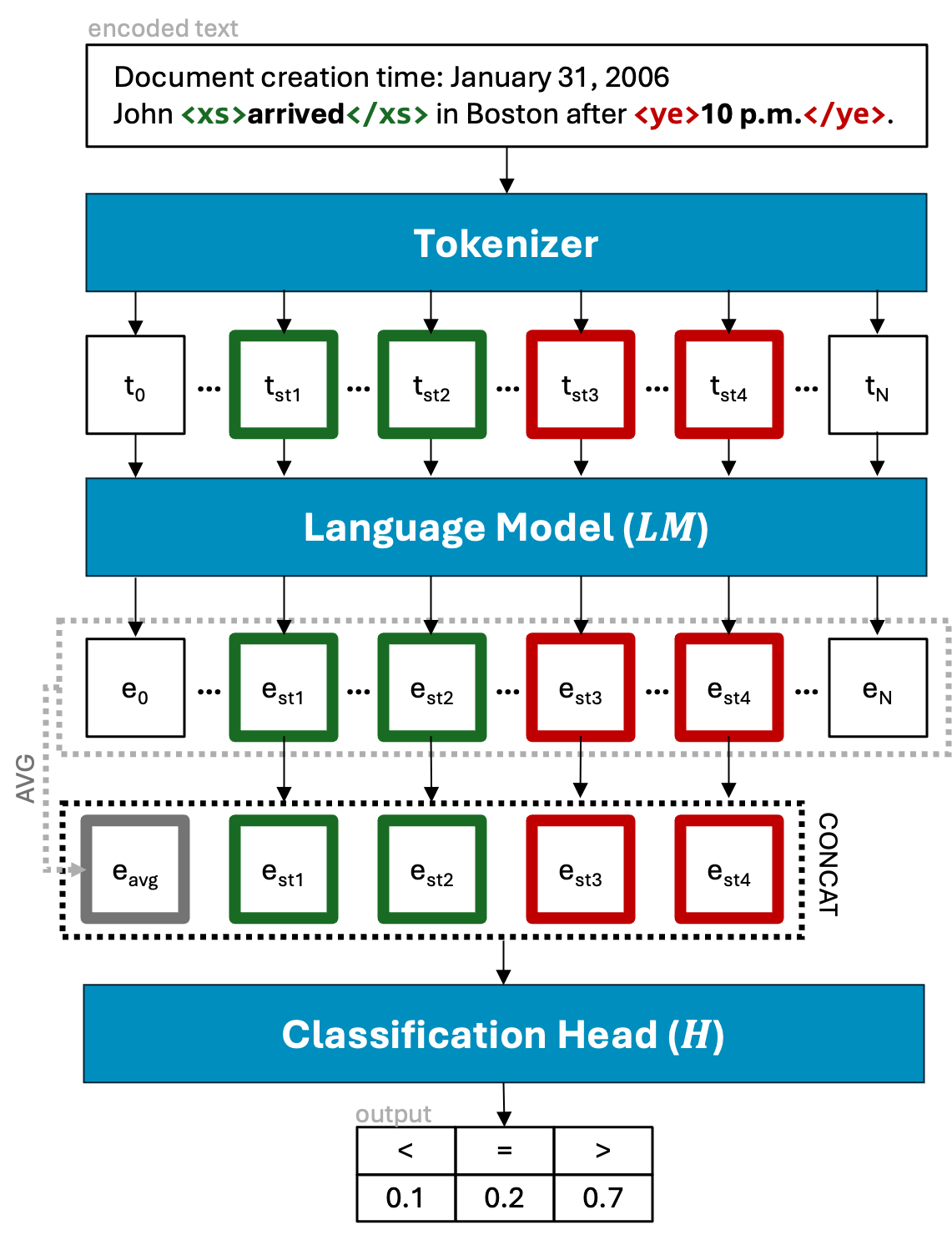}
    \caption{Point relation classification model classifying the relation between the start of \textit{arrived} and the end of \textit{10 p.m.}. The $t$ and $e$ in the diagram stand for tokens and embeddings, respectively.}
    \Description{Diagram of the point relation classification model. The input text \enquote{Document creation time: <dct> John <xs> arrived </xs> in Boston after <ye> 10 p.m. </ye>} is tokenized into tokens $t_1, \dots, t_N$, which are passed through a pre-trained language model that produces token embeddings $e_1, \dots, e_N$. The embeddings of the four special boundary tokens are concatenated with the average of all token embeddings to form a single vector that is fed to a feed-forward classification head, which outputs a probability distribution over the three point relations: before ($<$), equals ($=$), and after ($>$).}
    \label{fig:point-relation-classification}
\end{figure}

For the point relation classification we fine-tune a pre-trained language model $LM$ on a point relation dataset.
Before fine-tuning, we add eight tokens to its vocabulary and embedding table: \texttt{<xs>}, \texttt{</xs>}, \texttt{<xe>}, \texttt{</xe>}, \texttt{<ys>}, \texttt{</ys>}, \texttt{<ye>}, and \texttt{</ye>}.
These tokens are used as XML tags to mark the entities in the document and indicate the endpoint pair being classified.
Figure~\ref{fig:point-relation-classification} illustrates how the text \textit{\enquote{John \textbf{arrived} in Boston after \textbf{10 p.m.}}} would be tagged to classify the point relations between the start of \textit{\enquote{arrived}} and the end of \textit{\enquote{10 p.m.}}.
Since some  relations involve the document creation time as one of the entities, we prepend each document with the sentence: \enquote{Document creation time: \texttt{<dct>}}.
This allows us to treat the document creation time as any other entity.

The encoded document (with the XML tags and the document creation time) is processed by $LM$ to generate token embeddings $e_n\in\mathbb{R}^m$ for $n\in\{1,\dots,N\}$, where $m$ is the embedding dimension and $N$ the number of tokens.
Each encoded document contains four special tokens, marking entity boundaries at positions $st_k$ for $k\in\{1,2,3,4\}$.
We concatenate their embeddings, $e_{st}=(e_{st_1},e_{st_2},e_{st_3},e_{st_4})$, with the average token embedding, $e_{avg}=\sum_{n=1}^{N}e_n/N$, to form the vector $e=(e_{st},e_{avg})\in\mathbb{R}^{5m}$ to be used by the classification head $H$ which outputs the probability distribution over the three point relations $P(d,x_i,y_j)$:

\begin{equation}
    \begin{split}
        H(e) & = \sigma(W_2 \text{ReLU}(W_1(e) + b_1) + b_2) \\
             & = P(d, x_i, y_j)
    \end{split}
\end{equation}

where $W_1\in\mathbb{R}^{5m\times5m}$, $b_1\in\mathbb{R}^{5m}$, $W_2\in\mathbb{R}^{5m\times3}$, $b_2\in\mathbb{R}^3$, and $\sigma$ is the softmax function.

\subsection{Decoding Strategy} \label{ssec:decoding-strategy}

For the decoding strategy we use the estimated probability distributions $\mathbf{\hat{p}_{ij}}$ of the point relations to estimate the probabilities of the interval relations.
For each interval relation $r \in R$, we compute the probability $p_r$ as the product of the probabilities of the point relations that describe $r$:

\begin{equation}
    \hat{p}_r = \prod_{i \in \{s,e\}} \prod_{j \in \{s,e\}} \hat{p}_{ij,r_{ij}}
\end{equation}

where $\hat{p}_{ij,r_{ij}}$ is the estimated probability in Equation~(\ref{eq:point-relation-distribution}) and $r_{ij}$ is the point relation for the endpoint pair $(x_i,y_j)$ in $r$.
For instance, the predicted probability for the interval relation \texttt{starts} would be $\hat{p}_{starts} = \hat{p}_{ss,=} \times \hat{p}_{se,<} \times \hat{p}_{es,>} \times \hat{p}_{ee,<}$.
The predicted interval relation by the decoding strategy is then the one with the highest probability:

\begin{equation}
    \hat{r} = \arg \max_{r \in R} \hat{p}_r
\end{equation}

This is then used as the predicted interval relation by the IfP system.

\section{Experimental Setup} \label{sec:experimental-setup}

In this section, we describe the training and evaluation process for the point and interval relation classification systems.
To ensure reproducibility and encourage further research, we release the code for the experiments and the models that resulted from it under a permissive license\footnote{\url{https://github.com/hmosousa/temporal_classifier}}.

\subsection{Datasets} \label{ssec:datasets}

Since no dataset has been manually annotated with the four endpoint relations required to train the point relation model, we rely on existing interval-based datasets and convert their annotations into point relations.
However, this conversion imposes certain constraints.

In particular, we are restricted to datasets that provide fine-grained interval annotations to avoid introducing mislabeled point relations.
For instance, if the true relation between $x$ and $y$ is \texttt{meet}, but the dataset only provides a coarse label such as \texttt{before}, the conversion may incorrectly result in the point-level relation $(x_e, <, y_s)$ instead of the correct $(x_e, =, y_s)$.
As a consequence, widely used datasets such as TimeBank-Dense and TDDiscourse, which feature coarse-grained annotations, are unsuitable for evaluating our approach.

To the best of our knowledge, the only open dataset that satisfies the required annotation granularity is the TempEval-3~\cite{UzZaman2013} dataset\footnote{The MAVEN-ERE dataset~\cite{Wang2022} was also considered during our development. However, it was excluded because the test set labels are not publicly available, which limits comprehensive analysis of the results.}.
Released as part of the SemEval-13 campaign, it contains 275 news articles annotated according to the TimeML 1.2.1 guidelines~\cite{Saur2006}, which define 14 distinct temporal relation labels, namely: \texttt{BEFORE}, \texttt{AFTER}, \texttt{INCLUDES}, \texttt{IS\_INCLUDED}, \texttt{DURING}, \texttt{DURING\_INV}, \texttt{SIMULTANEOUS}, \texttt{IAFTER}, \texttt{IBEFORE}, \texttt{IDENTITY}, \texttt{BEGINS}, \texttt{ENDS}, \texttt{BEGUN\_BY}, and \texttt{ENDED\_BY}.
However, there is some confusion in the literature regarding these relations, as they conceptually overlap.
In particular, some inconsistencies exist regarding the interpretation of \texttt{DURING} and \texttt{DURING\_INV}.
The SemEval-13 competition paper does not mention \texttt{DURING\_INV}, and the dataset contains only a single occurrence of this label, suggesting a potential annotation anomaly.
Additionally, while~\citet{UzZaman2011} maps \texttt{DURING} to \texttt{IS\_INCLUDED}, the official evaluation script~\cite{UzZaman2013Toolkit} instead maps it to \texttt{SIMULTANEOUS}, aligning with its description in the TimeML 1.2.1 annotation guidelines.
Furthermore, the TempEval-3 paper lists \texttt{IDENTITY} as a label; however, the evaluation script maps it to \texttt{SIMULTANEOUS}.
To ensure comparability with the SemEval-13 results, we adopt the mapping used in their evaluation script which transforms the \texttt{DURING}, \texttt{DURING\_INV}, and \texttt{IDENTITY} relations to \texttt{SIMULTANEOUS}, as shown in Table~\ref{tab:timeml-to-allen-relations}.

\subsubsection{Data Augmentation} \label{sssec:data-augmentation}

Following~\citet{Mirza2014}, we augment the training set with two strategies:

\begin{itemize}
    \item \textbf{Inverse relations}: Since each temporal relation has an inverse, we can augment the dataset by adding the inverse relations. For example, ($x_s$, \texttt{<}, $y_s$) implies that ($y_s$, \texttt{>}, $x_s$). This doubles the dataset and may enhance generalization.
    \item \textbf{Temporal closure}: Using the transitivity property of temporal relations, we can infer additional relations from existing ones. For instance, if a document includes ($x_s$, \texttt{<}, $y_s$) and ($y_s$, \texttt{<}, $z_s$), we can infer that ($x_s$, \texttt{<}, $z_s$). This process, known as temporal closure~\cite{10.1145/182.358434}, is particularly advantageous at the point level, where more relations can be inferred than at the interval level. For example, if a document has the annotation $x$ \texttt{starts} $y$ and $x$ \texttt{starts} $z$, no interval relation can be inferred. However, at the point level, we can infer that $(y_s,\texttt{=},z_s)$, $(y_s,\texttt{<},z_e)$, and $(y_e,\texttt{>},z_s)$.
\end{itemize}

\begin{figure}[!ht]
    \centering
    \includegraphics[width=\columnwidth]{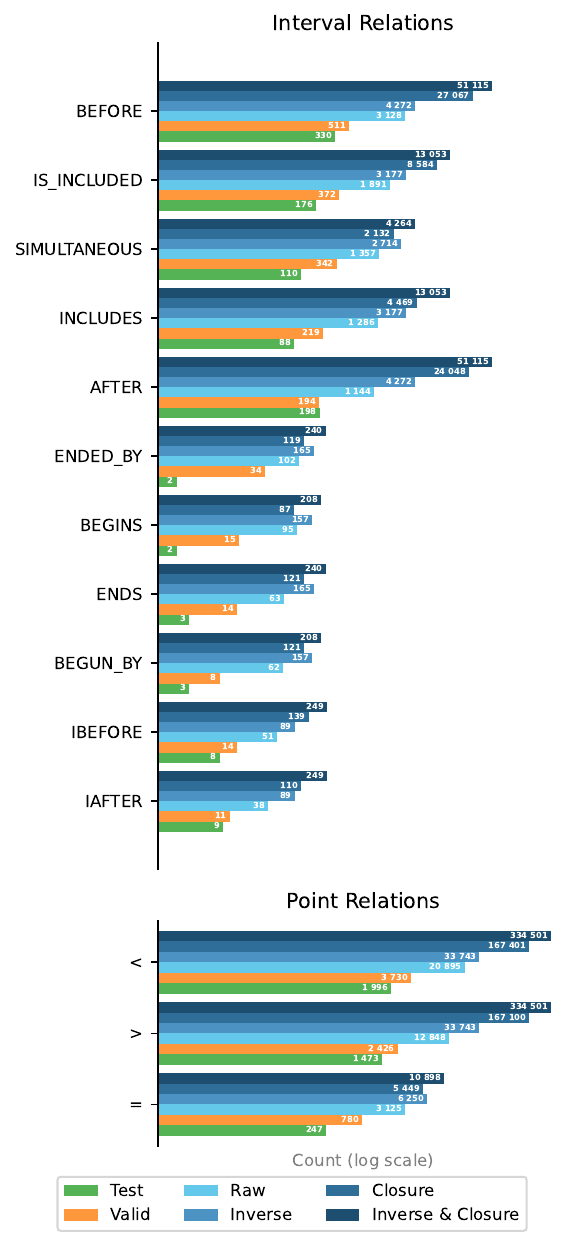}
    \caption{Interval and point datasets label distribution.}
    \Description{Stacked horizontal bar charts (with counts on a logarithmic scale) showing the label distribution for the interval relations and the point relations across the Test, Validation, Raw, Inverse, Closure, and Inverse \& Closure splits. The interval chart shows that BEFORE, IS\_INCLUDED, and SIMULTANEOUS dominate, while IBEFORE, IAFTER, BEGINS, BEGUN\_BY, ENDS, and ENDED\_BY are heavily underrepresented. The point chart shows that the $<$ and $>$ relations are far more frequent than $=$, with the imbalance growing as augmentation is applied.}
    \label{fig:dataset-overview}
\end{figure}

These augmentation strategies are used to generate the Inverse, Closure, and Inverse \& Closure training sets at the interval and point levels. Figure~\ref{fig:dataset-overview} shows the label distribution across these training sets, as well as for the training set without augmentation (Raw) and the validation and test sets.
The test set is the same as in the SemEval-13 competition, and the validation set consists of 20\% of the training documents.
It is important to note that the balance between \texttt{<} and \texttt{>} resulting from the closure transformation is artificial.
The point closure method in \texttt{tieval}~\cite{Sousa2023}, the Python library we used to infer point relations, outputs all relations as either \texttt{<} or \texttt{=}.
To address this constraint, we randomly replace half of the \texttt{<} relations with \texttt{>} relations.

\begin{table}[!ht]
    \centering
    \caption{The mapping of the TimeML relations to the Allen relations.}
    \label{tab:timeml-to-allen-relations}
    {\small
    \begin{tblr}{
        colspec = {X[c]X[c]},
        width=\columnwidth,
        row{1} = {c},
        column{1,2} = {c},
        vline{2,4} = {-}{0.05em},
        hline{1,2,16} = {-}{0.08em},
            }
        \textbf{TimeML}        & \textbf{Allen}        \\
        \texttt{BEFORE}        & \texttt{before}       \\
        \texttt{AFTER}         & \texttt{after}        \\
        \texttt{IBEFORE}       & \texttt{meet}         \\
        \texttt{IAFTER}        & \texttt{meet by}      \\
        \texttt{BEGINS}        & \texttt{start}        \\
        \texttt{BEGUN\_BY}     & \texttt{started by}   \\
        \texttt{ENDS}          & \texttt{finish}       \\
        \texttt{ENDED\_BY}     & \texttt{finished by}  \\
        \texttt{INCLUDES}      & \texttt{contains}     \\
        \texttt{IS\_INCLUDED}  & \texttt{during}       \\
        \texttt{DURING}        & \texttt{simultaneous} \\
        \texttt{DURING\_INV}   & \texttt{simultaneous} \\
        \texttt{SIMULTANEOUS } & \texttt{simultaneous} \\
        \texttt{IDENTITY}      & \texttt{simultaneous} \\
    \end{tblr}
    
    }
\end{table}

\subsection{Evaluation Metrics} \label{ssec:evaluation-metrics}

We evaluate both point and interval predictions using accuracy and the macro-average $F_1$ score over the label set.
For interval evaluation, we additionally report temporal awareness metric $F_{a}$~\citep{UzZaman2011}, which is an $F_1$ score that takes into account the temporal closure of annotated and predicted relations.
While this metric was tailored to assess the effectiveness of temporal extraction systems -- i.e., systems that identify and classify temporal relations -- prior work on temporal relation classification has also adopted them.
For a fair comparison temporal awareness is computed using the TempEval-3 evaluation script~\cite{UzZaman2013Toolkit}.
All results presented are for a single run of the model.

\subsection{Training Details} \label{ssec:training-details}
For the $LM$ we experimented with the SmolLM2-135M and SmolLM2-360M~\cite{Allal2025}.
Using these models allows us to evaluate the impact of (approximately) doubling the number of parameters on effectiveness.
Each model is fine-tuned in the four training sets presented in Section~\ref{ssec:datasets} to assess the benefits of the augmentation strategies. Furthermore, given their small size, we can perform hyperparameter optimization to mitigate the risk of conclusions being influenced by specific configurations.
The training is done by minimizing the cross-entropy loss and the search optimizes for macro-average $F_1$ score on the validation set, evaluating the model every 2,000 steps.
Training incorporates early stopping with a patience of 10 evaluation steps.
We use AdamW as the optimizer with a cosine learning rate schedule, with 2\% of the training steps being used for learning rate warm up.
The training runs were conducted on a machine with four NVIDIA L40S GPUs, each with 45GB of VRAM, and an AMD EPYC Milan processor with 28 cores and 220GB of RAM.

The hyperparameter search explores the following ranges: learning rate log-uniformly sampled between $1e-5$ and $1e-3$, number of training epochs uniformly sampled between $1$ and $10$, maximum gradient normalization uniformly sampled between $0.3$ and $1.0$, batch size from $\{4, 8, 16, 32\}$, and label smoothing uniformly sampled between $0.01$ and $0.1$.
Each (model, training set) undergoes 30 trials, with up to four trials running in parallel, each on one NVIDIA L40S GPU.
The search is conducted using Optuna~\cite{Akiba2019} with asynchronous successive halving~\cite{Li2020}.
Since our decoding strategy infers interval relations based on model-predicted confidence, proper calibration -- ensuring predicted probabilities align with actual accuracy -- is crucial.
Following~\citet{10.5555/3454287.3454709}, which have shown that label smoothing improves calibration, we apply label smoothing as a tunable hyperparameter.
Each hyperparameter search took on average 24 hours to complete which results in an estimate of 800 GPU hours to complete all training runs.
The best hyperparameters for each model are shown in Appendix~\ref{sec:hyperparameters}.

\subsection{Baseline Systems} \label{ssec:baseline-systems}

To ground our evaluation in easily replicable baselines, we report results for the \textbf{Random} system, which randomly samples a label at inference time, and the \textbf{Majority} system, which always predicts the most frequent label from the training set.

For the interval evaluation, we also compare our results with the most effective systems from previous studies. Specifically, we include \textbf{UTTime}~\cite{Laokulrat2013}, the most effective system in the SemEval-13 relation classification task, along with more recent models that achieved superior results: \textbf{Graph Stacking}~\cite{Laokulrat2015}, \textbf{TRelPro}~\cite{Mirza2014}, and \textbf{CATENA}~\cite{Mirza2016}. Additionally, we report results from \textbf{SP+ILP}~\cite{Ning2017}, which, unlike standard classification systems, incorporates a mechanism to ensure document-level timeline consistency. To the best of our knowledge, \textbf{SP+ILP} was the last work to publish classification results on TempEval-3.

To assess the baseline effectiveness of the $LM$s used on temporal relation classification, we also fine-tune them directly on the four interval training sets.
The architecture and text processing is the same as the one described in Section~\ref{ssec:point-relation-classification} but only four new tokens are added to the vocabulary: \texttt{<x>}, \texttt{</x>}, \texttt{<y>}, and \texttt{</y>}.
We refer to these models as \textbf{Interval-135M} and \textbf{Interval-360M}.

\section{Results \& Discussion} \label{sec:results-discussion}


\subsection{Point Results} \label{ssec:point-results}

\begin{figure}[!ht]
    \centering
    \includegraphics[width=\columnwidth]{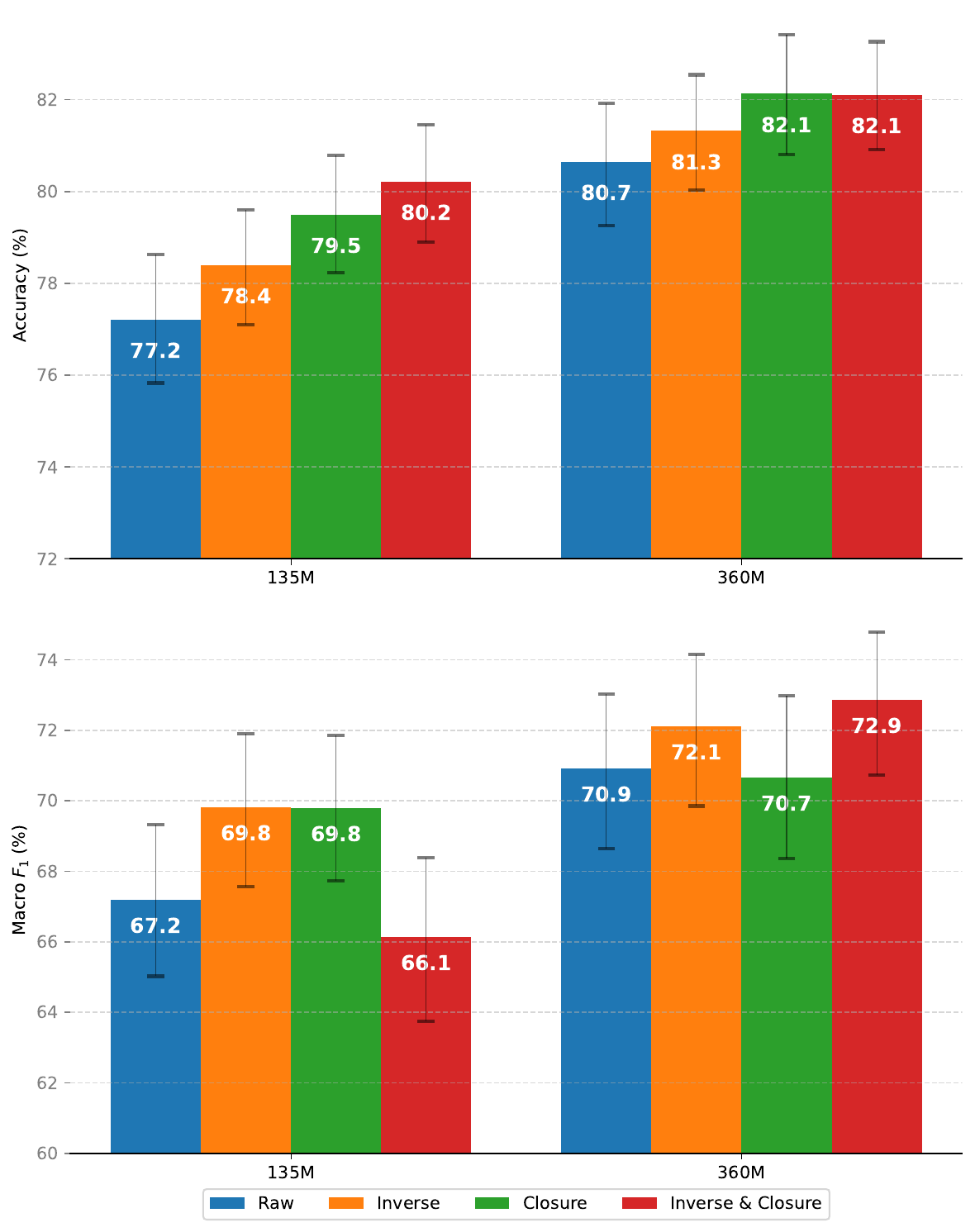}
    \caption{Accuracy and macro $F_1$ score of the point relation model that result from the different training sets. Error bars show the $95\%$ confidence interval, estimated using bootstrap with $1~000$ samples.} \label{fig:point-results}
    \Description{Two grouped bar charts comparing the 135M and 360M point models trained on the Raw, Inverse, Closure, and Inverse \& Closure datasets. The left chart reports accuracy (range 72--82\%) and the right chart reports macro $F_1$ (range 60--74\%). Both metrics increase from Raw to the augmented datasets, with the 360M model trained on the Inverse \& Closure dataset reaching the highest macro $F_1$ score of 72.9\%. Error bars indicate 95\% bootstrap confidence intervals.}
\end{figure}

Figure~\ref{fig:point-results} shows the accuracy and macro $F_1$ score of the point relation models trained on the four training sets: Raw (R), Inverse (I), Closure (C), and Closure \& Inverse (IC).
All models achieve an accuracy above $77\%$ and an $F_1$ score above $65\%$, significantly outperforming the Random and Majority baselines, which achieve $33.9\%$ and $53.7\%$ accuracy and $29.8\%$ and $23.3\%$ $F_1$ score, respectively.
The results reveal that for the 135M model, increasing the dataset size leads to a more accurate model. However, the same trend is not observed for the 360M model, where expanding the training data from the C to the IC dataset yields the same accuracy. However, when looking at the $F_1$ score, we see that the 360M model trained on the IC dataset achieves the highest $F_1$ score of all the models ($72.9\%$), indicating that it better balances effectiveness across all labels.

\begin{figure}[!ht]
    \centering
    \includegraphics[width=\columnwidth]{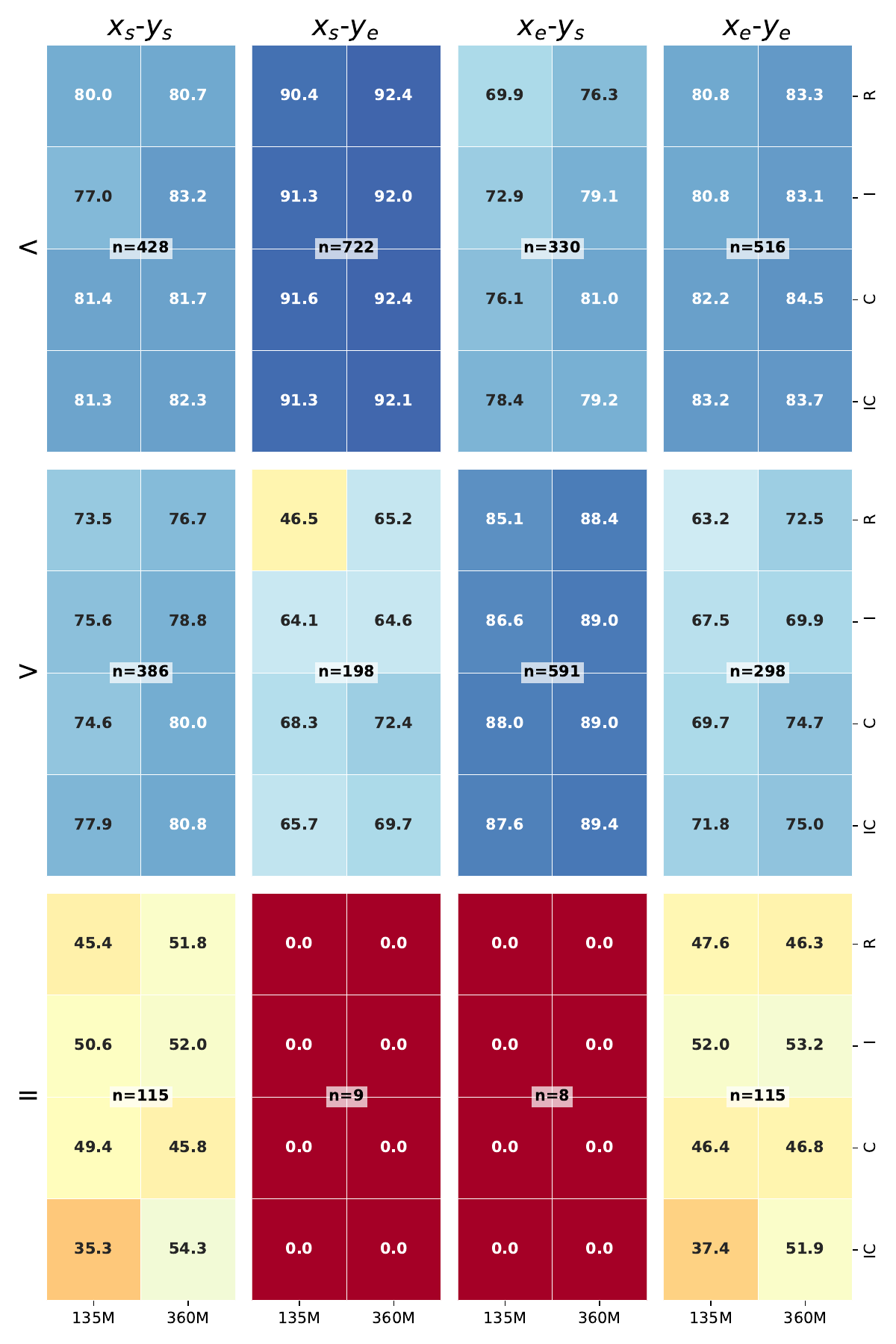}
    \caption{$F_1$ score of the point relation model for each (label, endpoint pair). $n$ is the support.} \label{fig:point-results-per-label}
    \Description{Heatmaps of per-label $F_1$ scores for the point relation model, broken down by the four endpoint pairs ($x_s$-$y_s$, $x_s$-$y_e$, $x_e$-$y_s$, $x_e$-$y_e$) and the three point relations ($<$, $=$, $>$). Each cell shows scores for the 135M and 360M models trained on the Raw, Inverse, Closure, and Inverse \& Closure datasets. The $<$ and $>$ relations achieve high $F_1$ (typically 70--92\%), the $=$ relation is much harder (35--75\% on $x_s$-$y_s$ and $x_e$-$y_e$, and 0\% on the underrepresented $x_s$-$y_e$ and $x_e$-$y_s$ pairs that have only 9 and 8 test instances).}
\end{figure}

When comparing models trained on the C dataset with the models trained on the I dataset, we can observe that the former achieve higher accuracy but a lower macro $F_1$ score.
This occurs because the C dataset is more imbalanced, leading to lower effectiveness on the \texttt{=} label.
This can be confirmed by looking at the last row of heatmaps in Figure~\ref{fig:point-results-per-label}, where it shows that the 360M model trained on the I dataset achieves an $F_1$ score of $52.0\%$ on the $(x_s, y_s)$ endpoint pair but only $45.8\%$ when trained on the C dataset.
The same effect can be observed in the $(x_e, y_e)$ endpoint pair and the 135M model.
On the other hand, the $F_1$ on the \texttt{<} and \texttt{>} relations increases when the model is trained on C rather than I.

Figure~\ref{fig:point-results-per-label} also highlights that the models struggle the most with the \texttt{=} relation, as its effectiveness on this label is lower than on \texttt{<} and \texttt{>}. However, it is important to note that this relation is underrepresented in the test set, particularly for the $(x_s,y_e)$ and $(x_e,y_s)$, with only 9 and 8 instances labeled for these pairs, respectively.
Therefore, future work should focus on improving the model's ability to learn the \texttt{=} relation but also on developing a benchmark where this relation is more represented.

Also in Figure~\ref{fig:point-results-per-label}, one can see that the models are highly effective at classifying $(x_s, y_e)$ as \texttt{<} and $(x_e,y_s)$ as \texttt{>}.
The effectiveness of this may be partially justified by the fact that these are the most represented relations in the training set, as shown in Table~\ref{tab:train_point_data_per_label}.
However, further research is needed to determine whether this is solely due to dataset imbalance or if these relations are inherently easier to learn for these endpoint pairs.

\begin{table}[ht]
    \centering
    \caption{Distribution of point relations by endpoint pair in the training sets. \textbf{R}, \textbf{I}, \textbf{C}, and \textbf{IC} refer to the Raw, Inverse, Closure, and Inverse \& Closure datasets, respectively.}
    \label{tab:train_point_data_per_label}
    { \small
        \begin{tblr}{
            colspec = {X[l, 1]X[c, 0.5]X[c, 2]X[c, 2]X[c, 2]X[c, 2]},
            width=\columnwidth,
            row{-} = {r},
            column{1} = {l},
            column{2} = {c},
            row{1} = {c},
            cell{2}{1} = {r=3}{},
            cell{5}{1} = {r=3}{},
            cell{8}{1} = {r=3}{},
            cell{11}{1} = {r=3}{},
            vline{2, 3} = {-}{0.05em},
            hline{1,2,14} = {-}{0.08em},
            hline{5,8,11,14} = {-}{},
                }
                        &   & $x_s-y_s$ & $x_s-y_e$ & $x_e-y_s$ & $x_e-y_e$ \\
            \textbf{R}  & $<$ & 4~567     & 8~035     & 3~128     & 5~165     \\
                        & $>$ & 3~136     & 1~144     & 6~038     & 2~530     \\
                        & $=$ & 1~514     & 38        & 51        & 1~522     \\
            \textbf{I}  & $<$ & 7~703     & 14~073    & 4~272     & 7~695     \\
                        & $>$ & 7~703     & 4~272     & 14~073    & 7~695     \\
                        & $=$ & 3~028     & 89        & 89        & 3~044     \\
            \textbf{C}  & $<$ & 15~406    & 28~146    & 8~544     & 15~390    \\
                        & $>$ & 15~406    & 8~544     & 28~146    & 15~390    \\
                        & $=$ & 6~056     & 178       & 178       & 6~088     \\
            \textbf{IC} & $<$ & 30~812    & 56~292    & 17~088    & 30~780    \\
                        & $>$ & 30~812    & 17~088    & 56~292    & 30~780    \\
                        & $=$ & 12~112    & 356       & 356       & 12~176    \\
        \end{tblr}
    }

\end{table}

\subsection{Interval Results} \label{ssec:interval-results}

\begin{figure*}[!ht]
    \centering
    \includegraphics[width=\textwidth]{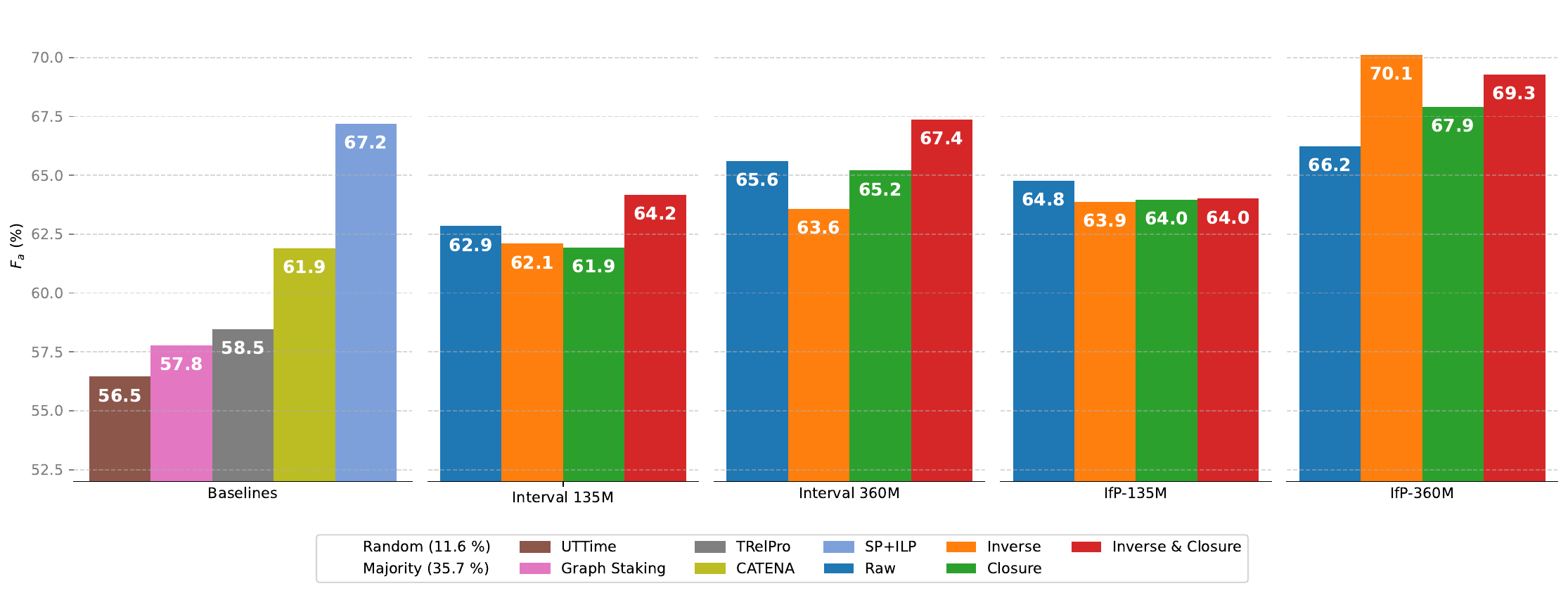}
    \caption{The temporal awareness of the baselines, the Interval models, and the IfP system on the TempEval-3 test set.} \label{fig:interval-results}
    \Description{Bar chart showing the temporal awareness $F_a$ score (\%) on the TempEval-3 test set across four groups of systems: Baselines (Random 11.6\%, Majority 35.7\%, UTTime 56.5\%, Graph Stacking 57.8\%, TRelPro 58.5\%, CATENA 61.9\%, SP+ILP 67.2\%), Interval-135M (62.9, 62.1, 61.9, 64.2), Interval-360M (65.6, 63.6, 65.2, 67.4), IfP-135M (64.8, 63.9, 64.0, 64.0), and IfP-360M (66.2, 70.1, 67.9, 69.3). For each model group the bars correspond to the four training sets: Raw, Inverse, Closure, and Inverse \& Closure. The IfP-360M trained on the Inverse dataset achieves the best score (70.1\%), surpassing all baseline systems.}
\end{figure*}

Figure~\ref{fig:interval-results} shows the $F_a$ score of the baselines, Interval models, and IfP systems.
Looking at the results of the Interval models, one can see that fine-tuning the 360M model on I yields the highest $F_a$ score, surpassing the best baseline system, SP+ILP.
However, the other Interval-360M models fail to achieve this level of effectiveness.
During training, we observed that all 360M models memorized the training data (i.e., achieved a training $F_a$ score of 1), while the 135M Interval models only memorized the R and I datasets.
This suggests that training the 360M model effectively requires either more data or regularization techniques to enhance generalization.

\begin{figure*}[!ht]
    \centering
    \includegraphics[width=\textwidth]{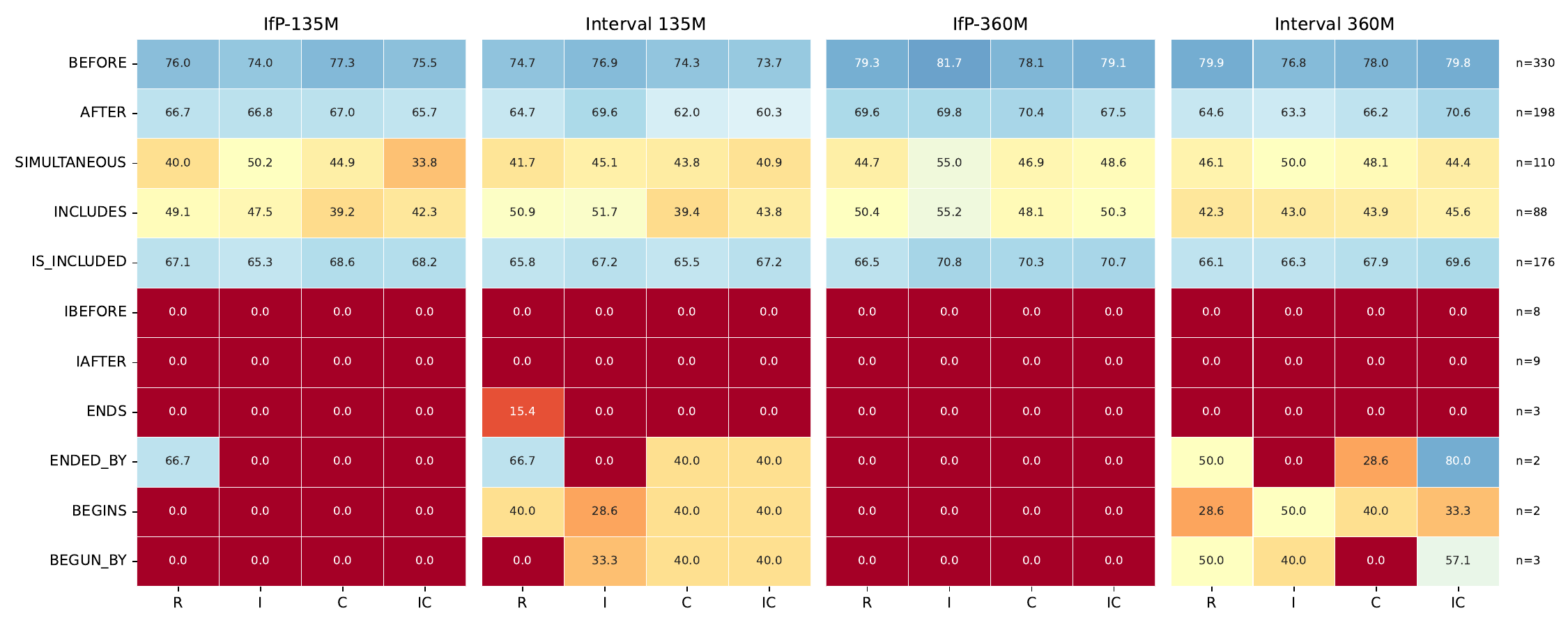}
    \caption{$F_1$ score of the Interval models and IfP systems for each label.}
    \Description{Four heatmaps reporting the per-label $F_1$ score (\%) of the Interval-135M, Interval-360M, IfP-135M, and IfP-360M systems trained on the Raw, Inverse, Closure, and Inverse \& Closure datasets. Rows are the 11 interval relations (BEFORE, AFTER, SIMULTANEOUS, INCLUDES, IS\_INCLUDED, IBEFORE, IAFTER, ENDS, ENDED\_BY, BEGINS, BEGUN\_BY) and columns are the four training sets. The IfP systems only successfully classify BEFORE, AFTER, SIMULTANEOUS, INCLUDES, and IS\_INCLUDED, scoring zero on the rarer labels, whereas the Interval models occasionally classify a few rarer labels. IfP-360M obtains the highest scores on SIMULTANEOUS and INCLUDES.}
    \label{fig:interval-relation-heatmaps}
\end{figure*}

\begin{figure*}[!ht]
    \centering
    \begin{subfigure}[b]{0.48\textwidth}
        \centering
        \includegraphics[width=\textwidth]{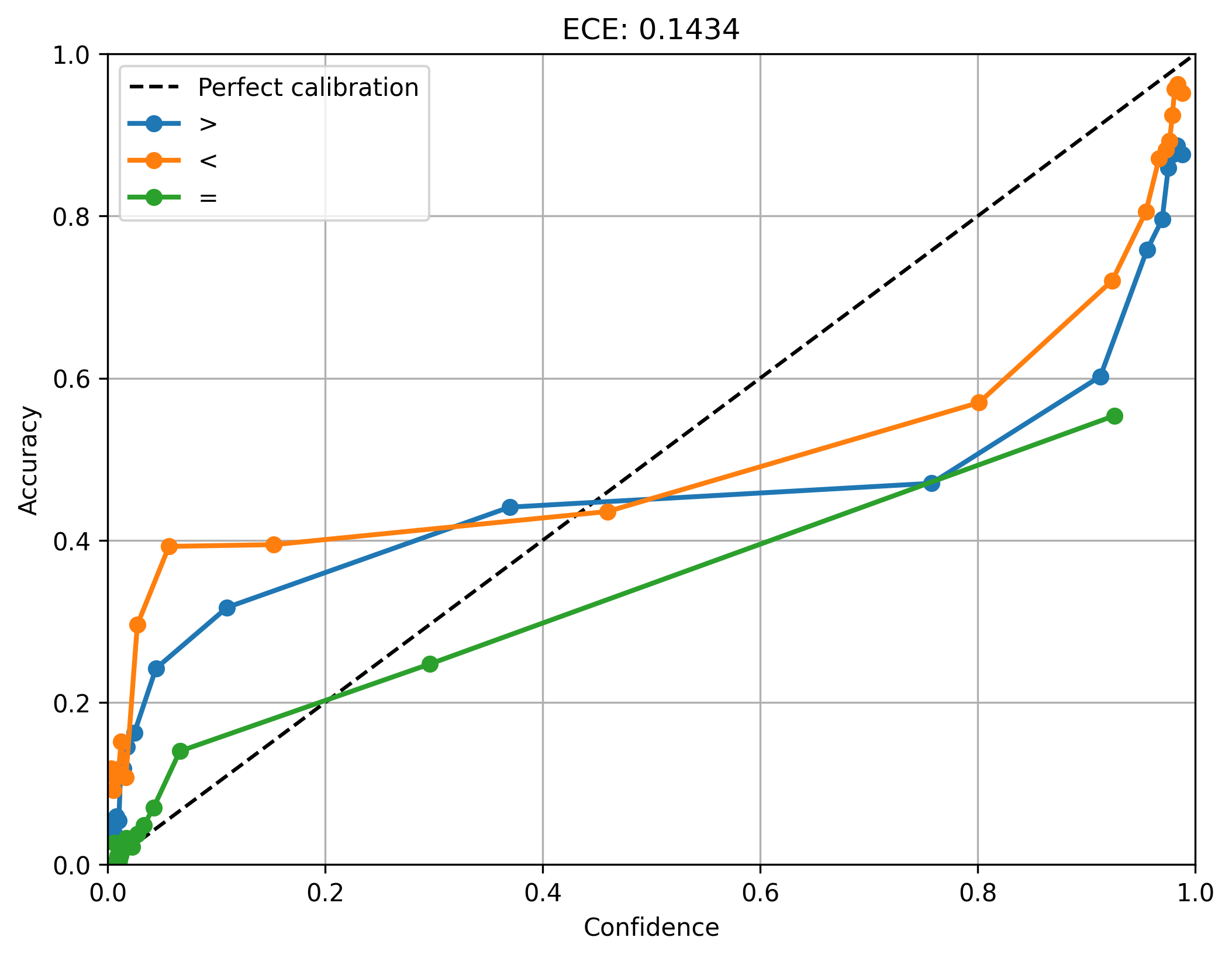}
        \caption{SmolLM2-135M trained on Inverse dataset calibration curve.}
        \Description{Calibration curve of the SmolLM2-135M point model trained on the Inverse dataset, plotted per label ($<$, $=$, $>$) on a one-vs-rest basis. The x-axis is the mean predicted confidence per quantile bin and the y-axis is the percentage of positive examples. The $<$ and $>$ curves cross the diagonal around 0.5: the model is overconfident below 0.5 and underconfident above 0.5. The $=$ curve lies above the diagonal across the whole range, indicating that the model is consistently underconfident on this label.}
        \label{fig:calibration-135}
    \end{subfigure}
    \hfill
    \begin{subfigure}[b]{0.48\textwidth}
        \centering
        \includegraphics[width=\textwidth]{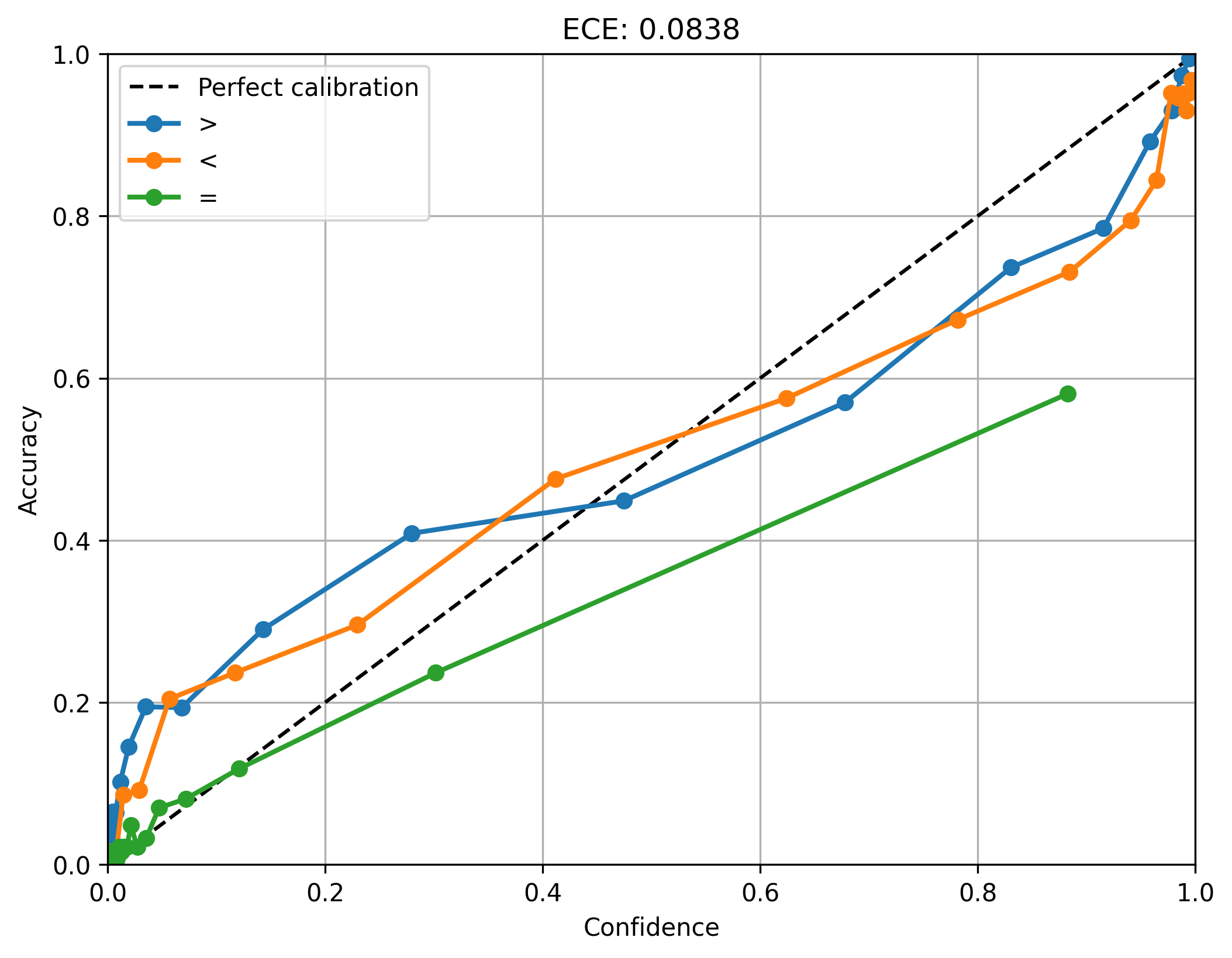}
        \caption{SmolLM2-360M trained on Inverse dataset calibration curve.}
        \Description{Calibration curve of the SmolLM2-360M point model trained on the Inverse dataset, plotted per label ($<$, $=$, $>$) on a one-vs-rest basis. The x-axis is the mean predicted confidence per quantile bin and the y-axis is the percentage of positive examples. The curves are closer to the diagonal than for the 135M model, reflecting a lower expected calibration error (8.38\% vs 14.3\% for the 135M model). The $=$ curve still lies above the diagonal, indicating residual underconfidence on the equals label.}
        \label{fig:calibration-360}
    \end{subfigure}
    \caption{Calibration curves of the model. The x-axis shows the mean predicted confidence, and the y-axis shows the percentage of positives. The curves are computed on a one-vs-rest basis using 20 bins over the quantile of the predicted confidence.}
    \label{fig:calibration}
\end{figure*}

For IfP, the model trained with 360M parameters achieved the highest $F_a$ score of $70.1\%$.
Interestingly, the point model underlying this model is neither the most accurate nor the most effective based on pointwise evaluation.
Comparing the results of Interval-360M and IfP-360M trained on I, we found that the improvement in effectiveness results from a higher $F_1$ score on the \texttt{SIMULTANEOUS} and \texttt{INCLUDES} relations, as shown in Figure~\ref{fig:interval-relation-heatmaps}.
The figure also shows that the IfP systems end up only classifying the temporal relations as \texttt{BEFORE}, \texttt{AFTER}, \texttt{SIMULTANEOUS}, \texttt{INCLUDES}, and \texttt{IS\_INCLUDED}.
This is a consequence of the low confidence that the model produces for the \texttt{=} relation, which can be observed in Figure~\ref{fig:calibration}.
The low probability makes any interval relation that has \texttt{=} in the point relations that describe it less likely by default.
Referring to Figure~\ref{fig:point-results-per-label}, we observe that the point model underlying IfP-360M trained on I has a high $F_1$ score on the \texttt{=} relation, which contributes to its high effectiveness on the \texttt{SIMULTANEOUS} relation.

For the \texttt{INCLUDES} relation, the point model achieves the highest effectiveness on the $(x_s, \texttt{<}, y_s)$ while having a good effectiveness on $(x_s, \texttt{<}, y_e)$ and $(x_e, \texttt{>}, y_s)$ -- three of the point relations that define \texttt{INCLUDES}.
This explains the model's high effectiveness on this relation.

Interestingly, training on I consistently improved the effectiveness over training on R, except in the case of the IfP-135M model.
Upon investigation, we found that the underlying point model of IfP-135M has an expected calibration error of $14.3\%$, whereas the IfP-360M model's point model has a significantly lower error of $8.38\%$. While the reason for the weaker calibration of the IfP-135M point model is unclear, this finding suggests that further research on point model calibration could enhance the overall effectiveness of the IfP system.

Since our decoding strategy uses the confidence of point relation predictions as a proxy for their probability, checking how well the model confidence aligns with the percentage of errors is critical.
Figure~\ref{fig:calibration} presents the calibration curve per label, computed on a one-vs-rest basis using 20 bins over the quantile of the predicted confidence.
From this figure, we can see that for the \texttt{<} and \texttt{>} labels the model is overconfident in its predictions when the confidence is below $0.5$ and underconfident when the confidence is above $0.5$.
For \texttt{=}, the model is underconfident for the whole range and has a tendency to predict lower confidence scores, since the average of the last quantile of the predicted confidence is around $0.7$.
This is a consequence of the few examples of the \texttt{=} label in the train and test sets.

In summary, our experiments show that fine-tuning a language model for temporal relation classification yields high effectiveness.
Additionally, using point representations as an intermediate step for predicting interval relations proves to be a successful strategy.
To further enhance the IfP system's effectiveness, future work should focus on three key areas: improving the representation of the \texttt{=} relation in the training set, ensuring a more balanced distribution of (source, target) pairs in the training set, and refining the calibration of the point model.

\section{Related Work} \label{sec:related-work}

Many approaches to temporal relation classification have been explored since the introduction of TimeBank~\cite{Pustejovsky2003TimeBank}.
Initially, the approaches compiled a feature set based on various types of semantic and linguistic features, which were then passed to either a rule-based system~\cite{Hagege2007}, a machine learning model~\cite{Mani2006}, or a hybrid system~\cite{Laokulrat2013,Souza2013,Mirza2014} to classify the temporal relations.
However, given that classifying relations independently can lead to temporally incoherent timelines at the document level, researchers have explored strategies to enforce temporal coherence at inference~\cite{Verhagen2008,Meng2017} and learning time~\cite{Ning2017,Leeuwenberg2017}.

With the advent of dense vector representation of text the field moved away from feature engineering, allowing neural models to learn representations directly from data.
Early neural approaches leveraged LSTM- and CNN-based architectures~\cite{Tourille2017,Dligach2017,Leeuwenberg2018}, while more recent work has built on pre-trained language models~\cite{Ning2019,Ross2020,Man2022,Wang2023}.
Various strategies have been explored to improve representation learning for temporal relation classification, including distant supervision~\cite{Ning2018c,Zhao2021,Zhou2021}, auxiliary training tasks~\cite{Ning2018b,Ballesteros2020,Wang2020}, and graph neural networks~\cite{Liu2021,Mathur2021,Zhang2022,Zhou2022}.

The work most closely related to ours is that of~\citet{Huang2023}, which also explores the use of point relations to infer interval relations.
However, their method focuses on event-event relation extraction (which also includes the vague label), relies on binary classification with logical mapping, and is evaluated only on TimeBank-Dense~\cite{Cassidy2014} and MATRES~\cite{Ning2018} -- limiting the analysis to a partial set of interval relations.
In contrast, we model point relation probabilities to infer the most likely interval relation, evaluate on the full set of interval relations, and include both events and temporal expressions as entities.
Furthermore, to the best of our knowledge, we are the first to explore the use of temporal closure at the point level.

\section{Conclusion \& Future Work} \label{sec:conclusion-future-work}

In this work, we propose a novel approach for temporal relation classification that predicts endpoint relations using a pointwise model and then decodes them into interval relations.
Our empirical evaluation on the TempEval-3 dataset shows the effectiveness of this method.
Despite these improvements, our analysis highlights the classification of the equal relation as a major bottleneck in the pointwise model.
One of the major limitations of this work is that the only suitable dataset to evaluate the system is the TempEval-3 dataset.
Future work should focus on creating a point-level dataset so that more extensive research can be conducted.
Furthermore, future work should also explore extending the system to document-level inference.



\begin{acks}
This work is funded by national funds through FCT - Fundação para a Ciência e a Tecnologia, I.P., under the support \texttt{\small UID/50014/2025} (\url{https://doi.org/10.54499/UID/50014/2025}).
\end{acks}

\appendix
\section{Hyperparameters} \label{sec:hyperparameters}

Table~\ref{tab:hyperparameters} shows the best hyperparameters for each model.

\begin{table}[ht]
    \centering
    \caption{The combination of hyperparameters that yielded the highest validation macro $F_1$ score for each model.}
    \label{tab:hyperparameters}
    \resizebox{\columnwidth}{!}{%
        \begin{tblr}{
            colspec = {l|c|c|c|c|c|c|c},
            width=\columnwidth,
            row{1-17} = {c},
            column{1} = {l},
            cell{1}{2} = {c=2}{},
            cell{1}{4} = {r=2}{},
            cell{1}{5} = {r=2}{},
            cell{1}{6} = {r=2}{},
            cell{1}{7} = {r=2}{},
            cell{1}{8} = {r=2}{},
            cell{3}{1} = {r=4}{},
            cell{7}{1} = {r=4}{},
            cell{11}{1} = {r=4}{},
            cell{15}{1} = {r=4}{},
            vline{2,5} = {-}{0.05em},
            hline{1,3,19} = {-}{0.08em},
            hline{7,11,15} = {-}{0.05em},
                }
                          & \textbf{Data} &            & \shortstack{Learning\\Rate} & \shortstack{Max\\Grad Norm} & \shortstack{\#~Epochs} & \shortstack{Label\\Smoothing} & \shortstack{Batch\\Size} \\
                          & I             & C          &               &               &           &                 &            \\
            Interval-135M &               &            & 0.0002        & 0.55          & 15        & 0.02            & 8          \\
                          & \checkmark    &            & 0.0003        & 0.60          & 25        & 0.03            & 4          \\
                          &               & \checkmark & 0.0002        & 0.40          & 12        & 0.02            & 16         \\
                          & \checkmark    & \checkmark & 0.0003        & 0.45          & 8         & 0.02            & 16         \\
            Interval-360M &               &            & 0.0001        & 0.70          & 10        & 0.02            & 4          \\
                          & \checkmark    &            & 0.0002        & 0.90          & 8         & 0.02            & 4          \\
                          &               & \checkmark & 0.0002        & 0.75          & 6         & 0.05            & 8          \\
                          & \checkmark    & \checkmark & 0.0003        & 0.45          & 5         & 0.02            & 8          \\
            Point-135M      &               &            & 0.0002        & 0.51          & 17        & 0.02            & 8          \\
                          & \checkmark    &            & 0.0003        & 0.61          & 29        & 0.03            & 4          \\
                          &               & \checkmark & 0.0002        & 0.36          & 10        & 0.02            & 16         \\
                          & \checkmark    & \checkmark & 0.0003        & 0.50          & 5         & 0.02            & 16         \\
            Point-360M    &               &            & 0.00009       & 0.67          & 12        & 0.02            & 4          \\
                          & \checkmark    &            & 0.0002        & 0.93          & 9         & 0.02            & 4          \\
                          &               & \checkmark & 0.0002        & 0.77          & 5         & 0.06            & 8          \\
                          & \checkmark    & \checkmark & 0.0003        & 0.41          & 5         & 0.02            & 8          \\
        \end{tblr}
    } 

\end{table}

\bibliographystyle{ACM-Reference-Format}
\bibliography{references}

\end{document}